\documentclass[conference]{IEEEtran}

\usepackage{cite}
\usepackage{amsmath,amssymb,amsfonts}
\usepackage{algorithmic}
\usepackage{graphicx}
\usepackage{textcomp}
\usepackage{xcolor}
\usepackage{subfiles}
\usepackage{multirow}
\usepackage[caption=false, font=footnotesize]{subfig}

\usepackage{assets/figures}
\usepackage{assets/equations}
\usepackage{assets/tables}

\def\BibTeX{{\rm B\kern-.05em{\sc i\kern-.025em b}\kern-.08em
    T\kern-.1667em\lower.7ex\hbox{E}\kern-.125emX}}

\begin{document}

\IEEEoverridecommandlockouts
\IEEEpubid{\makebox[\columnwidth]{978-1-6654-9810-4/22/\$31.00 \copyright 2022 IEEE \hfill} \hspace{\columnsep}\makebox[\columnwidth]{}}

\title{A Review and Implementation of Object Detection Models and Optimizations for Real-time Medical Mask Detection during the COVID-19 Pandemic\\
}

\author{\IEEEauthorblockN{Ioanna C. Gogou}
\IEEEauthorblockA{\textit{Dept. of Computer Engineering and Informatics} \\
\textit{University of Patras}\\
Patras, Greece \\
gogou@ceid.upatras.gr}
\and
\IEEEauthorblockN{Dimitrios A. Koutsomitropoulos}
\IEEEauthorblockA{\textit{Dept. of Computer Engineering and Informatics} \\
\textit{University of Patras}\\
Patras, Greece \\
koutsomi@ceid.upatras.gr}
}

\maketitle


\begin{abstract}
Convolutional Neural Networks (CNN) are commonly used for the problem of object detection thanks to their increased accuracy. Nevertheless, the performance of CNN-based detection models is ambiguous when detection speed is considered. To the best of our knowledge, there has not been sufficient evaluation of the available methods in terms of the speed/accuracy trade-off in related literature. This work assesses the most fundamental object detection models on the Common Objects in Context (COCO) dataset with respect to this trade-off, their memory consumption, and computational and storage cost. Next, we select a highly efficient model called YOLOv5 to train on the topical and unexplored dataset of human faces with medical masks, the  Properly-Wearing Masked Faces Dataset (PWMFD), and analyze the benefits of specific optimization techniques for real-time medical mask detection: transfer learning, data augmentations, and a Squeeze-and-Excitation attention mechanism. Using our findings in the context of the COVID-19 pandemic, we propose an optimized model based on YOLOv5s using transfer learning for the detection of correctly and incorrectly worn medical masks that surpassed more than two times in speed (69 frames per second) the state-of-the-art model SE-YOLOv3 on the PWMFD dataset while maintaining the same level of mean Average Precision (67\%).
\end{abstract}



\section{Introduction}
Computer vision has become an integral part of modern systems in transportation, manufacturing, and healthcare. In the last decade, the task of object detection as a deep learning problem has accumulated immense scientific interest. Convolutional Neural Networks (CNN) have shown excellent results in extracting the abstract features of image data, thanks to their similarities to the biological neural networks of the human brain \cite{cnn}. Their promising capabilities have motivated scientists towards brilliant inventions of new state-of-the-art object detectors resulting in a continuous increase in accuracy. Nevertheless, their performance is ambiguous when detection speed is considered, which is usually sacrificed in favor of accuracy. Conducting accurate object detection in real-time is a realistic requirement of modern systems, especially embedded ones with hardware limitations. However, the available methods have yet to be fully evaluated as published research \cite{review1,review2,tradeoff1,tradeoff2} tends to overlook the trade-off between accuracy and speed, compare models on different machine learning frameworks, or exclude newer models, resulting in indefinite results.

This work focuses on giving a solution to this problem by assessing the most fundamental CNN-based object detection models: Faster R-CNN \cite{fasterrcnn} and Mask R-CNN \cite{maskrcnn} of the family of Region-based Convolutional Neural Networks (R-CNN) \cite{rcnn}, RetinaNet \cite{retinanet}, Single Shot MultiBox Detector (SSD) \cite{ssd}, and You Only Look Once (YOLO) \cite{yolo} and its newer versions \cite{yolov2,yolov3,yolov4,yolov5}. The objective is to evaluate and compare them in terms of GPU memory consumption, computational and storage cost as well as their speed/accuracy trade-off. To reach fair conclusions, we execute the models through a common pipeline, using the same machine learning framework, dataset, and GPU. At the same time, our experiments can be reproduced through our accompanying open-source code.

Among those we evaluated, we choose a highly efficient model to train and optimize for real-time detection on a topical and novel dataset that has yet to be extensively tested. In view of the protective measures put in place during the COVID-19 pandemic, the need for real-time detection of correctly and incorrectly worn medical masks in data streams has become evident. According to the Worldwide Health Organization (WHO), the use of medical masks combined with other health measures is recommended for the containment of the virus \cite{who}. In this context, we propose an optimized real-time detector of correctly and incorrectly worn medical masks. To achieve top performance, we investigate optimization techniques used before in medical mask detection. For instance, transfer learning from larger and more diverse object detection datasets is expected to improve model accuracy.

The main contributions of this work are the following:
\begin{itemize}

    \item An analysis of the speed/accuracy trade-off of known object detectors using the same framework, dataset, and GPU. No other similar study has been published that includes YOLOv5\cite{yolov5}, whose performance has yet to be extensively tested.
    
    \item The accuracy and speed of YOLOv5s were evaluated for the first time on the newly-developed Properly-Wearing Masked Faces Dataset (PWMFD) \cite{seyolov3} dataset. Furthermore, the effects of transfer learning, data augmentations, and an attention mechanism were assessed for medical mask detection.
    
    \item A real-time medical mask detection model based on YOLOv5 was proposed that surpassed more than two times in speed (69 fps) the state-of-the-art model SE-YOLOv3 \cite{seyolov3} on the PWMFD dataset while maintaining the same level of mean Average Precision (mAP) at 67\%. This increase in speed gives room for using the model on embedded devices with lower hardware capabilities, while still achieving real-time detection.
    
    
    \item All results are reproducible through our open-source code on GitHub\footnote{ github.com/joangog/object-detection}. The weight file of our medical mask detector is also available on GitHub\footnote{github.com/joangog/object-detection-assets} and on Hugging Face\footnote{huggingface.co/joangog/pwmfd-yolov5}. 
    
\end{itemize}


\section{Previous Work}
\subsection{Object Detection Models}

The development of AlexNet \cite{alexnet} in 2012 paved the way towards the CNN-based object detection models we know today. Introduced in 2014, R-CNN \cite{rcnn} was the first one to adopt the idea of region proposals for object detection, produced through a selective search algorithm. In 2015, its successor, Fast R-CNN \cite{fastrcnn}, increased detection speed by computing a feature map for the whole image rather than for each proposal. It was improved further with Faster R-CNN \cite{fasterrcnn} by replacing selective search with a more efficient fully convolutional region proposal network. In 2017, Mask R-CNN \cite{maskrcnn} was an extension of Faster R-CNN for image segmentation.

In 2015, the first one-shot model was published named YOLO \cite{yolo}.  Two iterations followed, YOLOv2 \cite{yolov2} and YOLOv3 \cite{yolov3}, improving its performance with the addition of anchors, batch normalization, the Darknet-53 backbone, and three detection heads. In 2020, its development resumed with YOLOv4 \cite{yolov4}. It included the enhanced CSPDarknet-53 backbone, a Spatial Pyramid Pooling (SPP) \cite{spp} layer, and a Path Aggregation Network (PANet) \cite{panet}. Shortly after, YOLOv5 \cite{yolov5} was launched with only small alterations. Its role as the fifth version of YOLO was a controversial subject as no official publication has been released to this day. Nevertheless, it shows promising results through consistent updates, which are worth investigating. Finally, SSD \cite{ssd} proposed in 2016 and RetinaNet \cite{retinanet} in 2017 are two other notable one-shot models. The latter became known for introducing focal loss to combat foreground-background imbalance.

\subsection{Speed\slash Accuracy Trade-off of Object Detection Models}

In recent years, several reviews of modern object detection models have been published. In 2019, a survey \cite{review1} was conducted on the improvements in object detection in the last 20 years. Nonetheless, the survey merely reported on the performance of the models in question in related bibliography. No experiment was performed to measure their accuracy and speed. In contrast, another review in the same year \cite{review2} included an experimental evaluation of more than 26 models on the Visual Object Classes (VOC) \cite{voc} and Common Objects in Context (COCO) \cite{coco} datasets. Although both accuracy and speed were measured, the trade-off between the two parameters was not analyzed. Moreover, the tested models were implemented in different machine learning frameworks and programming languages, thus obscuring detection speed.

Two studies published in 2017 \cite{tradeoff1} and 2018 \cite{tradeoff2} assessed the speed/accuracy trade-off of object detection models using the same framework. In  \cite{tradeoff1}, the Tensorflow implementations of Faster R-CNN \cite{fasterrcnn}, SSD \cite{ssd}, and Region-based Fully Convolutional Network (R-FCN) \cite{rfcn} were evaluated on COCO \cite{coco} based on their speed/accuracy trade-off while testing different backbones, image resolutions, and numbers of region proposals. In \cite{ tradeoff2}, a similar analysis was conducted using the same models with the addition of Mask R-CNN \cite{maskrcnn} and SSDlite \cite{ssdlite}. The aspect of memory consumption and detection speed on different devices was investigated as well. Nevertheless, both studies do not take into account newer models, such as YOLOv5 \cite{yolov5} and RetinaNet \cite{retinanet}.

\subsection{Medical Mask Detection}

Initially, interest in medical mask detection was limited. After the outbreak of COVID-19, numerous medical mask detection models were proposed to limit infection.

In 2020, the Super-Resolution and Classification Network \cite{srcnet} was designed and trained using transfer learning. In 2021, RetinaFaceMask \cite{retinafacemask} was introduced, a one-shot model based on RetinaNet,  using transfer learning from a human face dataset to the Masked Faces for Face Mask Detection Dataset (MAFA-FMD) made by the authors. In the same year, a hybrid medical mask recognition model \cite{hybridmask} combining ResNet-50 \cite{resnet} with a Support Vector Machine (SVM) was proposed, after being trained on both real-world and synthetic data using transfer learning. Later in \cite{yolov2mask}, the authors replaced the SVM with YOLOv2 \cite{yolov2}. A medical mask detector published in \cite{inceptionv3mask} was based on Inception-v3 \cite{inceptionv3} and trained on a synthetic mask dataset using transfer learning from a general object dataset and several data augmentations.

Despite achieving near-perfect accuracy, the above models were not evaluated on their detection speed. In view of this, a real-time mask detection model was designed, SE-YOLOv3 \cite{seyolov3}, and trained on the novel Properly-Wearing Masked Face Detection (PWMFD) dataset created by the authors. It introduced a Squeeze-and-Excitation (SE) attention mechanism \cite{squeeze} to YOLOv3, achieving  66\% mAP and 28 fps. However, its performance was evaluated using a high-end GPU, rendering it unsuitable for lightweight devices. Moving to YOLOv4, the hybrid mask detection model tiny-YOLOv4-SPP was proposed\cite{tinyyolov4sppmask}. It significantly reduced training time while increasing accuracy compared to the original tiny-YOLOv4, but the aspect of real-time detection was not considered.

\section{Evaluation of Object Detection Models}\label{sec:evaluation}
\subsection{Evaluation Methodology}

We evaluated the object detection models shown in Table \ref{tab:modeltable}  in terms of memory consumption, computational and storage cost as well as their speed/accuracy trade-off. The models include Faster R-CNN with two different backbones and image sizes, Mask R-CNN, RetinaNet, SSD and its memory-efficient version SSDlite, YOLOv3 with its SPP and tiny variants, YOLOv4, and YOLOv5 including its large, medium, small, and nano variants. For all models, we utilized the same framework, dataset, and GPU.

\subsubsection{Framework}
All models are implemented in PyTorch and are offered in the Torchvision package. The only exceptions are YOLOv3, YOLOv4, and YOLOv5, which are implemented in GitHub repositories\footnote{github.com/ultralytics/yolov3}\footnote{github.com/Tianxiaomo/pytorch-YOLOv4}\footnote{github.com/ultralytics/yolov5}. 

\subsubsection{Dataset}
The evaluation is done on the val subset of the COCO2017 \cite{coco} dataset. COCO contains 118,000 examples for training and 5,000 for validation belonging to 80 classes of everyday objects. No training phase is performed as all models are pre-trained on the train subset. The data are loaded with a batch size of 1 to imitate the stream-like insertion when detecting in real-time.

\subsubsection{Environment}
The code for the experiment is organized in two Jupyter notebooks (coco17\_inference.pynb, analysis.pynb) and is executed through the Google Colab platform. We chose the option of a local runtime, which uses the GPU of our system (Nvidia Geforce GTX 960, 4 GB). Every model goes through the same inference pipeline.

\subsubsection{Metrics}
To estimate the memory usage of each model we calculate the maximum GPU memory allocated to our GPU device by CUDA for the program. To quantify computational costs, Giga Floating Point Operations (GFLOPs) are counted using the ptflops\footnote{github.com/sovrasov/flops-counter.pytorch} Python package. The storage cost is derived from the size of the weight file of each respective model. We measure detection speed in frames (images) per second (fps). For accuracy, we use the mean Average Precision (mAP) of all classes. According to the COCO evaluation standard\footnote{cocodataset.org/\#detection-eval}, Average Precision (AP) is calculated using 101-point interpolation on the area under the P-R (Precision-Recall) curve, as follows: 

\modeltable
\averageprecision

\subsection{Memory Consumption}

In Fig. \ref{fig:memory}, the higher the number of parameters and image size, the more GPU memory the model consumes. Interestingly, YOLOv4 has 925 MB of memory consumption,  three times more than models with roughly the same parameter count and image size, such as YOLOv3 (301 MB). In contrast, YOLOv3 utilizes approximately three-fourths of the memory consumed by YOLOv5l, despite having more parameters and the same image size. SSDlite, having the fewest parameters and smallest image size, uses merely 33 MB.

\subsection{Computational and Storage Cost}

Firstly, the number of GFLOPs of a model is closely related to its number of parameters and image size. For instance, in Fig. \ref{fig:gflops}, between the two implementations of Faster R-CNN with MobileNet-v3 and ResNet-50, the second executed 27 times more GFLOPs than the first while having just over twice the same number of parameters. In general, models that use the MobileNet-v3 backbone, i.e., SSDlite and Faster R-CNN, proved to be the least expensive in GFLOPs. Subsequently, comparing the two versions of Faster R-CNN with different image sizes, the one with size 800 cost almost six times more GFLOPs than the one with 320. The sole exception to the rule was RetinaNet and YOLOv3 that, despite having the same image size and fewer parameters than Faster R-CNN ResNet-50 and YOLOv4 respectively, executed more GFLOPs than their counterpart.



According to Fig. \ref{fig:storage}, storage cost increases linearly to the number of parameters in a model. Nevertheless, YOLOv3 and YOLOv5 cost significantly less storage space than models with an equivalent number of parameters.

\memorygflops
\storage

\subsection{Speed\slash Accuracy Trade-off}

In Fig. \ref{fig:mapfps} we illustrate the trade-off between the speed (fps) and accuracy (mAP) of all models. An immediate observation is a decrease in speed as the accuracy of a model increases. On a general note, a favorable model would be one that achieves both high accuracy and speed. Thus, it would be found in the top right corner of Fig. \ref{fig:mapfps}. All variations of YOLOv5 provided the best balance between accuracy and speed, whereas RetinaNet, SSD, SSDlite, and YOLOv3-tiny ranked last. 

\mapfps



\section{Implementation of Medical Mask Detector}
\subsection{Configuration}

\subsubsection{Dataset}
To achieve desirable results, a realistic and diverse dataset for both localization and recognition of medical masks is required; one that is large enough to ensure high accuracy but does not exceed our hardware limitations. Therefore, we selected the newly-created PWMFD \cite{seyolov3} dataset, using its train and validation subsets to train and evaluate our model respectively. It includes 9,205 real-life images with 18,532 annotations of faces belonging to three classes, “with mask”, “incorrect mask” and “without mask.”

\subsubsection{Environment}
The code for the experiment is executed through Google Colab and is organized in three Jupyter notebooks (mask\_training.pynb, mask\_inference.pynb, analysis.pynb). Training is performed using the GPU provided by Colab (Nvidia Tesla K80, 12 GB) due to its memory facilitating a larger batch size, whereas evaluation is performed using our local GPU (Nvidia Geforce GTX 960, 4 GB) because of its higher detection speed.

\subsubsection{Training and Evaluation}
During training, a batch size of 32 is used. Training lasts for 50 epochs, as more resulted in overfitting. Learning rate is updated according to the OneCycleLR \cite{onecyclelr} policy with values in the range of [0.001, 0.01]. As an optimizer, Stochastic Gradient Descent is employed with a value of 0.937 for momentum and 0.0005 for weight decay. Cross-Entropy is used for classification loss and Complete Intersection over Union (CIoU) for localization loss. After training, the final weights are selected from the epoch with the highest mAP. During evaluation, inference is performed with a batch size of 1 to mimic the sequential input of data in real-time. COCO mAP and fps are measured as metrics.

\subsection{Model}

According to our study in Section \ref{sec:evaluation}, YOLOv5 \cite{yolov5}, and particularly its medium, small and nano variants, provided the best balance between accuracy, speed, memory consumption, and computational and storage costs for real-time object detection. Its architecture is depicted in Fig. \ref{fig:yolovfive}. To implement our medical mask detector, we chose to train its small variant, YOLOv5s, on PWMFD \cite{seyolov3} with an image size of 320, achieving 33\% mAP and 69 fps. To elevate its performance, we experimented with various optimizations during training.

\yolovfive

\subsection{Optimizations}

\subsubsection{Transfer Learning}
Inspired by the success of transfer learning in previous medical mask detectors \cite{retinafacemask,srcnet,hybridmask,yolov2mask,inceptionv3mask}, we apply weights pre-trained on COCO \cite{coco} to PWMFD \cite{seyolov3}. This technique is known for significantly decreasing training time and the need for a large dataset \cite{transferlearning}, both crucial in our case. We tested various training schemes for YOLOv5s on PWMFD: without transfer learning using random initial weights (row 1 in Table \ref{tab:masktransfer}), and with transfer learning using the COCO weights (rows 2-4 in Table \ref{tab:masktransfer}), while experimenting with freezing the weights of different layers before training. The highest mAP (38\%) was achieved by freezing the pre-trained backbone, that is, training only the head on PWMFD.

\subsubsection{Data Augmentations}
To prevent overfitting, we utilize the following basic data augmentations: translation, scaling, flipping, and Hue-Saturation-Value (HSV) transformations. Furthermore, we assess the effect of two novel transformations, mosaic and mixup \cite{mixup}. The first combines multiple images forming a mosaic, while the second stacks two images on top of one another with a degree of transparency. The potential benefits of mosaic and mixup for YOLO were explored in \cite{yolov4} and in \cite{seyolov3} for mask detection. We trained YOLOv5s with three different data augmentation combinations as shown in Table \ref{tab:maskaugment}. Mosaic nearly doubled accuracy (mAP 67\%), but the addition of mixup was not beneficial.

\subsubsection{Attention Mechanism}
In \cite{seyolov3}, by introducing two SE blocks to the backbone of YOLOv3 as an attention mechanism along with mixup and focal loss, the accuracy of YOLOv3 on PWMFD rose by 8.6\%. The SE mechanism applies input-dependent weights to the channels of the feature map to create a better representation of the image. Focal loss is an improvement on Cross-Entropy loss that assists the model in focusing on hard misclassified examples during training. We apply the same strategy to YOLOv5. Our ablation study in Table \ref{tab:maskattention} shows that these optimizations did not affect speed but they impacted accuracy negatively. Therefore, they are not used in our final mask detector. Nevertheless, Fig. \ref{fig:maskattentionsize} illustrates higher accuracy for small and medium-sized objects when using SE with mixup.

\subsection{Evaluation}

Our final mask detector is based on YOLOv5s with transfer learning from COCO to PWMFD while freezing the backbone, and uses mosaic and other basic data augmentations. According to Table \ref{tab:maskfinal}, it is twice as accurate as the baseline YOLOv5s, while being equal in speed. At the same time, when compared using PWMFD, it is as accurate and more than two times faster on our own lower-end GTX 960 GPU than the state-of-the-art SE-YOLOv3 on a GTX 2070. This significant increase in speed gives room for its use in embedded devices with lower hardware capabilities, while still achieving real-time detection. Detection examples are illustrated in Fig. \ref{fig:maskexamples}.

\masktransfer
\maskaugment
\maskattention
\maskattentionsize
\maskfinal
\maskexamples

\section{Conclusion and Future Work}
Evaluating object detection models fairly is a task that requires multiple parameters to be addressed besides accuracy. Our goal is to provide an informative analysis of fundamental object detectors that also includes speed, memory consumption, and computational and storage costs. Using our findings, we propose an optimized YOLOv5s-based model for real-time mask detection to protect public health amidst the COVID-19 pandemic. Our optimizations led to increased accuracy (mAP 67\%) that rivaled that of state-of-the-art SE-YOLOv3 on PWMFD, while being more than two times faster (69 fps). At the same time, applying the SE attention mechanism of SE-YOLOv3 to YOLOv5s along with mixup improved accuracy for small and medium-sized objects, but not large ones.

In the future, we would like to research optimizations to combat the side effects on large objects, thus improving total accuracy. Also, our model does not utilize an important characteristic of data streams, the relationship between consecutive frames. This could be achieved by exploring object detection models that implement this using Recurrent Neural Networks.

After the end of the pandemic, we are hopeful that with the help of our model the healthcare sector can be better prepared for a similar crisis. Our proposed optimizations may also be useful for other related problems such as face detection. 

\bibliography{IEEEabrv,assets/bibliography}

\begin{thebibliography}{10}
\providecommand{\url}[1]{#1}
\csname url@samestyle\endcsname
\providecommand{\newblock}{\relax}
\providecommand{\bibinfo}[2]{#2}
\providecommand{\BIBentrySTDinterwordspacing}{\spaceskip=0pt\relax}
\providecommand{\BIBentryALTinterwordstretchfactor}{4}
\providecommand{\BIBentryALTinterwordspacing}{\spaceskip=\fontdimen2\font plus
\BIBentryALTinterwordstretchfactor\fontdimen3\font minus
  \fontdimen4\font\relax}
\providecommand{\BIBforeignlanguage}[2]{{%
\expandafter\ifx\csname l@#1\endcsname\relax
\typeout{** WARNING: IEEEtran.bst: No hyphenation pattern has been}%
\typeout{** loaded for the language `#1'. Using the pattern for}%
\typeout{** the default language instead.}%
\else
\language=\csname l@#1\endcsname
\fi
#2}}
\providecommand{\BIBdecl}{\relax}
\BIBdecl

\bibitem{cnn}
Y.~Lecun and Y.~Bengio, ``Convolutional networks for images, speech and time
  series,'' in \emph{The Handbook of Brain Theory and Neural Networks}.\hskip
  1em plus 0.5em minus 0.4em\relax The MIT Press, 1995.

\bibitem{review1}
Z.~Zou, Z.~Shi, Y.~Guo, and J.~Ye, ``Object detection in 20 years: {A}
  survey,'' \emph{arXiv preprint arXiv:1905.05055}, 2019.

\bibitem{review2}
Z.~Zhao, P.~Zheng, S.~Xu, and X.~Wu, ``Object detection with deep learning: {A}
  review,'' \emph{IEEE Trans. Neural Netw. Learn. Syst.}, vol.~30, no.~11, pp.
  3212--3232, 2019.

\bibitem{tradeoff1}
J.~Huang \emph{et~al.}, ``Speed/accuracy trade-offs for modern convolutional
  object detectors,'' in \emph{Proc. IEEE Conf. Comput. Vis. Pattern
  Recognit.}, 2017, pp. 3296--3297.

\bibitem{tradeoff2}
A.~Srivastava \emph{et~al.}, ``Performance and memory trade-offs of deep
  learning object detection in fast streaming high-definition images,'' in
  \emph{Proc. IEEE Int. Conf. Big Data}, 2018, pp. 3915--3924.

\bibitem{fasterrcnn}
S.~Ren, K.~He, R.~B. Girshick, and J.~Sun, ``Faster {R-CNN}: Towards real-time
  object detection with region proposal networks,'' in \emph{Proc. Adv. Neural
  Inf. Process. Syst.}, vol.~28, 2015.

\bibitem{maskrcnn}
K.~He, G.~Gkioxari, P.~Dollár, and R.~B. Girshick, ``Mask {R-CNN},'' in
  \emph{Proc. IEEE Int. Conf. Comput. Vis.}, 2017, pp. 2961--2969.

\bibitem{rcnn}
R.~B. Girshick, J.~Donahue, T.~Darrell, and J.~Malik, ``Rich feature
  hierarchies for accurate object detection and semantic segmentation,'' in
  \emph{Proc. IEEE Conf. Comput. Vis. Pattern Recognit.}, 2014, pp. 580--587.

\bibitem{retinanet}
T.~Lin, P.~Goyal, R.~B. Girshick, K.~He, and P.~Dollár, ``Focal loss for dense
  object detection,'' in \emph{Proc. IEEE Int. Conf. Comput. Vis.}, 2017.

\bibitem{ssd}
W.~Liu \emph{et~al.}, ``{SSD}: {S}ingle {S}hot {M}ultibox {D}etector,'' in
  \emph{Proc. Eur. Conf. Comput. Vis.}, 2016, p. 21–37.

\bibitem{yolo}
J.~Redmon, K.~Divvala, R.~B. Girshick, and A.~Farhadi, ``You only look once:
  Unified, real-time object detection,'' in \emph{Proc. IEEE Conf. Comput. Vis.
  Pattern Recognit.}, 2015, pp. 779--788.

\bibitem{yolov2}
J.~Redmon and A.~Farhadi, ``{YOLO9000:} better, faster, stronger,'' in
  \emph{Proc. IEEE Conf. Comput. Vis. Pattern Recognit.}, 2017, pp. 7263--7271.

\bibitem{yolov3}
------, ``{YOLOv3}: An incremental improvement,'' \emph{arXiv preprint
  arXiv:1804.02767}, 2018.

\bibitem{yolov4}
A.~Bochkovskiy, C.~Wang, and H.~M. Liao, ``{YOLOv4}: Optimal speed and accuracy
  of object detection,'' \emph{arXiv preprint arXiv:2004.10934}, 2020.

\bibitem{yolov5}
G.~Jocher \emph{et~al.}, \emph{{ultralytics/yolov5: v6.0 - YOLOv5n 'Nano'
  models, Roboflow integration, TensorFlow export, OpenCV DNN support}}, 2021.

\bibitem{who}
{World Health Organization}, ``Advice on the use of masks in the context of
  {COVID-19}: interim guidance, 6 april 2020,'' Tech. Rep., 2020.

\bibitem{seyolov3}
X.~Jiang, T.~Gao, Z.~Zhu, and Y.~Zhao, ``Real-time face mask detection method
  based on {YOLOv3},'' \emph{Electronics}, vol.~10, no.~7, 2021.

\bibitem{alexnet}
A.~Krizhevsky, I.~Sutskever, and G.~E. Hinton, ``Image{N}et classification with
  deep convolutional neural networks,'' in \emph{Proc. Adv. Neural Inf.
  Process. Syst.}, 2012, pp. 1097--1105.

\bibitem{fastrcnn}
R.~B. Girshick, ``Fast {R-CNN},'' in \emph{Proc. IEEE Int. Conf. Comput. Vis.},
  2015, pp. 1440--1448.

\bibitem{spp}
K.~He, X.~Zhang, S.~Ren, and J.~Sun, ``Spatial pyramid pooling in deep
  convolutional networks for visual recognition,'' \emph{IEEE Trans. Pattern
  Anal. Mach. Intell.}, vol.~37, no.~9, pp. 1904--1916, 2015.

\bibitem{panet}
S.~Liu, L.~Qi, H.~Qin, J.~Shi, and J.~Jia, ``Path aggregation network for
  instance segmentation.'' in \emph{Proc. IEEE Conf. Comput. Vis. Pattern
  Recognit.}, 2018, pp. 8759--8768.

\bibitem{voc}
M.~Everingham, L.~V. Gool, C.~K.~I. Williams, J.~M. Winn, and A.~Zisserman,
  ``The {P}ascal {V}isual {O}bject {C}lasses ({VOC}) challenge.'' \emph{Int. J.
  Comput. Vis.}, vol.~88, no.~2, pp. 303--338, 2010.

\bibitem{coco}
T.~Lin \emph{et~al.}, ``Microsoft {COCO}: {C}ommon {O}bjects in {C}ontext,'' in
  \emph{Proc. Eur. Conf. Comput. Vis.}, 2014, pp. 740--755.

\bibitem{rfcn}
J.~Dai, Y.~Li, K.~He, and J.~Sun, ``{R-FCN:} {O}bject detection via
  region-based fully convolutional networks,'' in \emph{Proc. Adv. Neural Inf.
  Process. Syst.}, vol.~29, 2016.

\bibitem{ssdlite}
M.~Sandler, A.~Howard, M.~Zhu, A.~Zhmoginov, and L.~Chen, ``Mobile{N}et{V}2:
  {I}nverted residuals and linear bottlenecks,'' in \emph{Proc. IEEE Conf.
  Comput. Vis. Pattern Recognit.}, 2018.

\bibitem{srcnet}
B.~Qin and D.~Li, ``Identifying facemask-wearing condition using image
  super-resolution with classification network to prevent {COVID-19},''
  \emph{Sensors}, vol.~20, no.~18, 2020.

\bibitem{retinafacemask}
X.~Fan and M.~Jiang, ``Retina{F}ace{M}ask: A single stage face mask detector
  for assisting control of the {COVID}-19 pandemic,'' in \emph{Proc. IEEE Int.
  Conf. Syst. Man Cybern. Syst.}, 2021, pp. 832--837.

\bibitem{hybridmask}
M.~Loey, G.~Manogaran, M.~H.~N. Taha, and N.~E.~M. Khalifa, ``A hybrid deep
  transfer learning model with machine learning methods for face mask detection
  in the era of the {COVID-19} pandemic,'' \emph{Measurement}, vol. 167, 2021.

\bibitem{resnet}
K.~He, X.~Zhang, S.~Ren, and J.~Sun, ``Deep residual learning for image
  recognition,'' in \emph{Proc. IEEE Conf. Comput. Vis. Pattern Recognit.},
  2016, pp. 770--778.

\bibitem{yolov2mask}
M.~Loey, G.~Manogaran, M.~H.~N. Taha, and N.~E.~M. Khalifa, ``Fighting against
  {COVID-19}: A novel deep learning model based on {YOLO}-v2 with {R}es{N}et-50
  for medical face mask detection,'' \emph{Sustainable Cities and Society},
  vol.~65, 2021.

\bibitem{inceptionv3mask}
G.~Jignesh~Chowdary, N.~S. Punn, S.~K. Sonbhadra, and S.~Agarwal, ``Face mask
  detection using transfer learning of {I}nception{V}3,'' in \emph{Proc. Int.
  Conf. Big Data Analytics}, 2020, p. 81–90.

\bibitem{inceptionv3}
C.~Szegedy, V.~Vanhoucke, S.~Ioffe, J.~Shlens, and Z.~Wojna, ``Rethinking the
  inception architecture for computer vision,'' in \emph{Proc. IEEE Conf.
  Comput. Vis. Pattern Recognit.}, 2016.

\bibitem{squeeze}
J.~Hu, L.~Shen, and G.~Sun, ``Squeeze-and-excitation networks,'' in \emph{Proc.
  IEEE Conf. Comput. Vis. Pattern Recognit.}, 2018.

\bibitem{tinyyolov4sppmask}
A.~Kumar, A.~Kalia, A.~Sharma, and M.~Kaushal, ``A hybrid tiny {YOLO} v4-{SPP}
  module based improved face mask detection vision system,'' \emph{J. Ambient
  Intell. Human. Comput.}, 2021.

\bibitem{mobilenetv3}
A.~Howard \emph{et~al.}, ``Searching for mobilenetv3,'' in \emph{Proc. IEEE
  Int. Conf. Comput. Vis.}, 2019.

\bibitem{vgg16}
K.~Simonyan and A.~Zisserman, ``Very deep convolutional networks for
  large-scale image recognition,'' \emph{arXiv preprint arXiv:1409.1556}, 2014.

\bibitem{onecyclelr}
L.~N. Smith and N.~Topin, ``Super-convergence: Very fast training of neural
  networks using large learning rates,'' in \emph{Proc. Artif. Intell. and
  Mach. Learn. for Multi-Domain Oper. Appl.}, 2019, pp. 369--386.

\bibitem{transferlearning}
C.~Tan, F.~Sun, T.~Kong, W.~Zhang, C.~Yang, and C.~Liu, ``A survey on deep
  transfer learning,'' in \emph{Proc. Int. Conf. Artif. Neural Netw.}, 2018,
  pp. 270--279.

\bibitem{mixup}
H.~Zhang, M.~Ciss{\'{e}}, Y.~N. Dauphin, and D.~Lopez{-}Paz, ``mixup: Beyond
  empirical risk minimization,'' \emph{arXiv preprint arXiv:1710.09412}, 2017.

\end{thebibliography}

\end{document}